\documentclass[lettersize,journal]{IEEEtran}
\usepackage{amsmath,amsfonts}
\usepackage{algorithmic}
\usepackage{algorithm}
\usepackage{array}
\usepackage[caption=false,font=normalsize,labelfont=sf,textfont=sf]{subfig}
\usepackage{textcomp}
\usepackage{stfloats}
\usepackage{url}
\usepackage{verbatim}
\usepackage{graphicx}
\usepackage{cite}
\usepackage{booktabs}
\usepackage{amssymb}
\usepackage[dvipsnames]{xcolor}
\begin{document}

\title{Balancing Multimodal Learning through Label Space Reshaping}

\author{Xiaoyu Ma, Weijie Zhang, Yuanhao Gao, Han Miao, Yongjian Deng, Hao Chen
\thanks{}}

\markboth{Preprint 2026.05}%
{Shell \MakeLowercase{\textit{et al.}}: Balancing Multimodal Learning through Label Space Reshaping}

\IEEEpubid{0000--0000/00\$00.00~\copyright~2021 IEEE}

\maketitle

\begin{abstract}
Multimodal learning often suffers from modality imbalance, where modalities that converge faster dominate optimization while others remain undertrained.
Existing approaches typically mitigate this issue by strengthening the weak modality or adjusting optimization gradients.
However, such strategies mainly compensate for optimization-rate discrepancies, often at the expense of the strong modality’s optimization capacity, without analyzing how these discrepancies arise at the modality level.
Based on theoretical insights and empirical observations, we argue that the discrepancy of learning pace arises from differences in the mapping difficulty between modality-specific feature space and the shared label space. 
To address this issue, we propose Balanced Multimodal Label Reshaping (BMLR), the first method that promotes multimodal balance from the label-side design. 
BMLR reshapes the cross-modal label space to equalize mapping difficulty across modalities, thereby facilitating modality interaction and injecting richer inter-class information into each modality.
Extensive experiments across multiple architectures demonstrate that BMLR consistently improves multimodal performance and exhibits strong compatibility with diverse model designs.
The source code will be released soon.
\end{abstract}


\section{Introduction}\label{sec:introduction}
\IEEEPARstart{M}{ultimodal} learning \cite{multimodal,multimodal2} aims to integrate information from diverse modalities \cite{vision+}, to achieve comprehensive perception and reasoning in complex scenarios.
Despite the advantages of cross-modal complementarity \cite{mmfusion}, studies have identified the problem of \textit{Modality Laziness} \cite{makes, laziness}, where some modalities fail to learn effectively due to \textit{Modality Imbalance} \cite{ogm}.
During joint training, models often exhibit a \textit{Greedy} \cite{greedy} tendency to optimize the strong modality while neglecting the weak modality, resulting in suboptimal multimodal learning.

\begin{figure}[!t]
  \centering
  \includegraphics[width=0.48\textwidth]{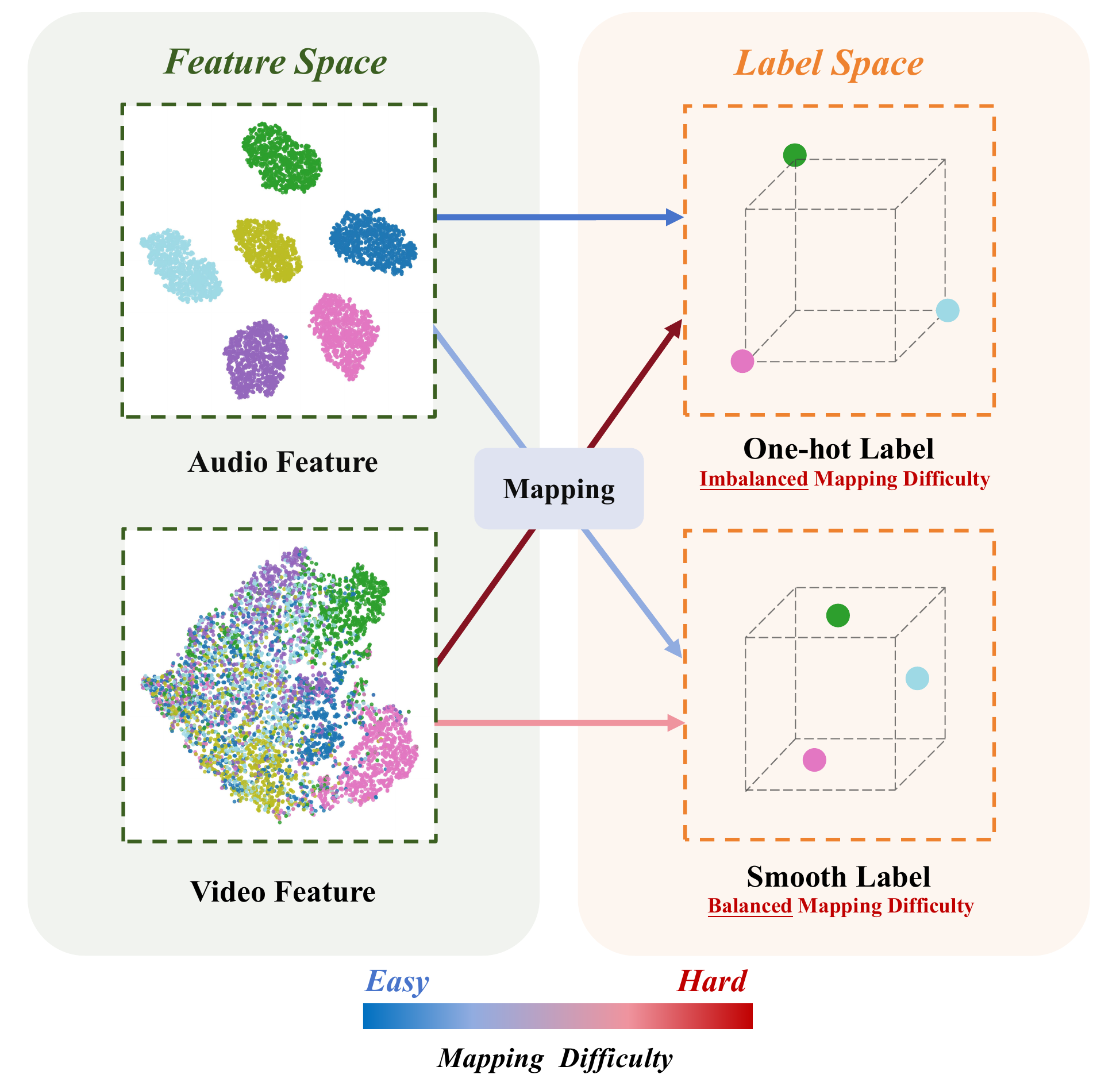}
  \caption{\textbf{Mapping relationship between feature space and label space.} The \textit{feature space} is derived from the CREMAD and visualized by t-SNE. The \textit{label space} is illustrated using three classes from CREMAD as examples.}
  \label{fig: motivation}
\end{figure}

Existing studies in \textbf{Balanced Multimodal Learning (BML)} \cite{bml} emphasize enhancing weak modalities to achieve balance and fully leverage multimodal inputs.
One research direction focuses on the optimization objective, designing modality-specific learning goals \cite{makes, pmr, mmcosine, uni-t, mmpareto} to strengthen the weak modality.
Representative techniques include refining task supervision \cite{makes}, incorporating prototype-based constraints \cite{pmr}, and leveraging knowledge distillation \cite{uni-t} to mitigate modality laziness.
Another direction targets optimization dynamics, adaptively modifying learning rates~\cite{ogm, agm, atf} or gradient updates~\cite{mgm, cgm} based on unimodal performance, aligning the optimization pace across modalities and preventing training bias.
In summary, while existing studies observe inconsistencies in modality optimization pace and attempt to compensate for them through supervision or optimization adjustments, they do not analyze the causes of these discrepancies at modality level or propose solutions that directly address them.

We argue that \textbf{the inconsistency of learning rates can be explained from the label side, as it stems from differences in the difficulty of mapping between the feature space and the label space.}
In multimodal classification, models are typically supervised with one-hot labels, whose discrete and discontinuous nature forms isolated orthogonal points in the label space \cite{onehot}. 
\IEEEpubidadjcol
This sharp label distribution creates a highly nonlinear optimization landscape, forcing the model to learn steep decision boundaries near class margins \cite{devil}. 
As shown in Figure \ref{fig: motivation}, different modalities exhibit distinct feature characteristics; those with clearer and more discriminative features can fit the label space more easily, while modalities with noisy or ambiguous semantics face greater optimization difficulty.
Prior work also supports this insight.
LFM \cite{lfm} noted the impact of one-hot labels on modality imbalance, and mitigated this issue by introducing contrastive learning to weaken label dependency, rather than tackling it in the label space.
It has been theoretically shown in \cite{understanding} that modalities better aligned with the target distribution are optimized earlier. 
We further extend this perspective by analyzing the mapping difficulty between the feature and label space across modalities in real-world settings.
This motivates us to address modality imbalance directly from the label side, which we consider a more fundamental and effective approach, as one-hot labels cannot fully capture all the information of the data inputs \cite{ldl}. 
By reshaping the label space, we can \textbf{align learning rates across modalities}, \textbf{incorporate inter-class information}, and \textbf{foster cross-modal interactions}, leading to further performance improvements.

\begin{figure*}[!t]
  \centering
  \includegraphics[width=0.99\textwidth]{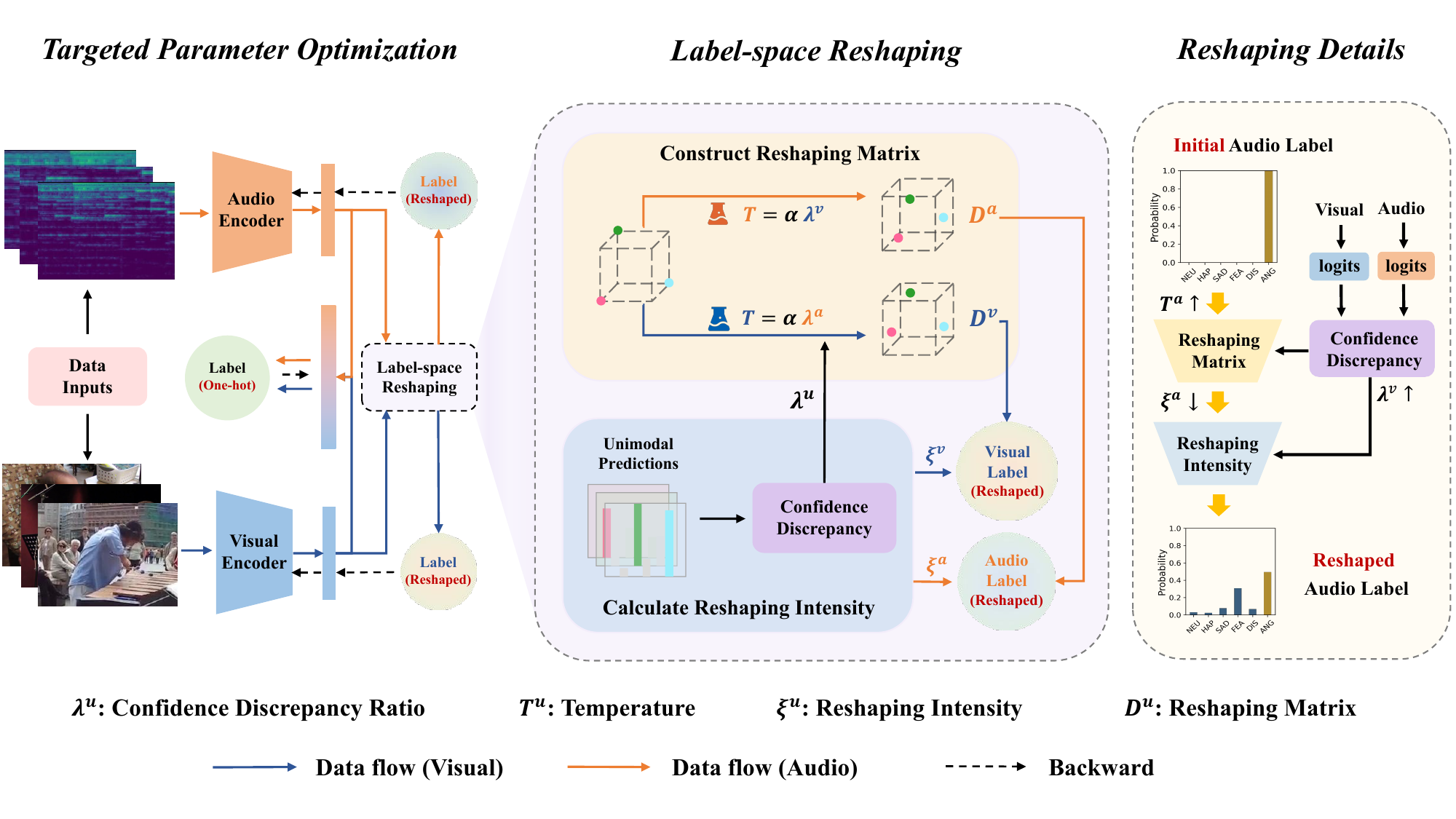}
  \caption{\textbf{Pipeline of the BMLR method.}
  We construct a balanced and class-relation-aware label space by leveraging cross-modal prediction comparison and interaction, while optimizing different parts of the model with modality-specific objectives.}
  \label{fig:method}
  \vskip -0.1in
\end{figure*}

Building on this motivation, we propose \textbf{Balanced Multimodal Label Reshaping (BMLR)}, which leverages unimodal outputs and confidence discrepancies to construct balanced and class-aware label spaces for each modality, thereby enabling more equitable multimodal learning, as illustrated in Figure \ref{fig:method}.
Concretely, we first perform cross-modal label-space reshaping based on the unimodal predictions of each modality. 
The \textit{Reshaping Matrix} is customized using predictions from another modality, which determines the structure of the label-space transformation, while the \textit{Reshaping Intensity} is governed by the confidence discrepancy among modalities, controlling the extent of the transformation. 
This reshaped label space not only aligns learning rates across modalities but also enriches inter-class relationships through cross-modal interactions.
To mitigate learning instability caused by dynamic label-space variations, we adopt a \textit{Targeted Parameter Optimization} strategy, applying different optimization objectives to different parts of the model. This ensures simultaneous improvement of unimodal representation learning and multimodal decision-making.
Our label-side balanced learning method aligns learning pace across modalities and enhances model capability by constructing a \textit{balanced and class-relation-aware label space} at the sample level. 
To summarize, our contributions in this paper are as follows:
\begin{itemize}
    \item We empirically analyze the underlying cause of the differences in modality optimization rates, which arise from varying mapping difficulties between the feature space and the label space.
    \item To the best of our knowledge, we propose the first label-side method to alleviate modality imbalance. 
    BMLR constructs a balanced, class-relation-aware label space that aligns optimization rates across modalities while providing richer cross-modal inter-class information.
    \item Experiments on multiple datasets, model architectures, and fusion methods demonstrate that BMLR consistently achieves significant performance gains, confirming its strong generalizability and effectiveness.
\end{itemize}

\section{Related Work}
\subsection{Balanced Multimodal Learning}
When training multimodal models with a unified objective, modality imbalance causes the strong modality to dominate optimization, leading to insufficient learning of others \cite{makes}. 

To address the imbalance, some methods design modality-specific objectives to strengthen the weak modality \cite{makes, uni-t, pmr, lfm, balancevl}. 
Other works adjust training process to balance optimization paces \cite{ atf, ogm, agm, opm, reconboost, amss, gradient}. 
To resolve the conflicts in training, some methods decouple the multimodal learning \cite{mla, remix}or adjust the optimization directions \cite{mmpareto}.
While these methods help align learning rates, they overlook the cause of varying learning difficulty \cite{understanding}.
We argue that such differences stem from the complexity of mapping between feature and label spaces, motivating our approach to align learning difficulty by reconstructing the label space.

Since our method is designed from the label side, each modality is required to learn relatively independent and dynamic supervision signals.
Inspired by DGL \cite{dgl}, which decouples encoder gradients from decision-layer gradients to avoid optimization conflicts, we optimize each modality encoder with its own unimodal objective, while the multimodal prediction loss is used to update only the decision layer.

\subsection{Label Smoothing and Knowledge Distillation}
\textit{Label Smoothing (LS)} is a regularization technique that replaces hard labels with a mixture of hard labels and a uniform distribution over all classes, preventing the model from becoming overconfident \cite{EarliesrSmooth, OutputDistribution}.
\textit{Knowledge Distillation (KD)} is a knowledge transfer paradigm in which a compact student model is trained to mimic a more accurate teacher model \cite{abkd, DistillSurvey,kdva}.
The core insight of KD is that the output probabilities of a trained model contain rich knowledge beyond hard labels.  
Extending this concept, KD can be viewed as a special form of LS, constructing soft labels that capture more realistic inter-class relationships \cite{smoothDistill}.
In this work, we build on this idea by constructing a more balanced, class-relationship-aware label space for each modality, guided by the modality-specific learning dynamics.
Importantly, our approach fundamentally differs from distillation-based balancing methods such as UMT \cite{uni-t}, which rely on intra-modal knowledge transfer, cannot facilitate cross-modal interaction, and require additional resources to pretrain unimodal models.

\section{Method}
\subsection{Model Formulation}
We focus on the multimodal discrimination task, following related works \cite{bml}. 
For simplicity, we consider an input scenario with two modalities: $m_{a}$ and $m_{v}$. 
The training dataset is $\mathcal{D} = \{\boldsymbol{x}_i, \boldsymbol{y}_i\}_{i=1, 2, \dots, N}$, where each $\boldsymbol{x}_{i}$ consists of multimodal inputs, i.e., $\boldsymbol{x}_{i} = (\boldsymbol{x}^{a}_{i},\boldsymbol{x}^{v}_{i})$, and $\boldsymbol{y}_{i} \in \{0, 1\}^C$, with $C$ denoting the number of classes. 
We employ a multimodal model comprising two dedicated unimodal encoders followed by a fusion network. Each unimodal encoder processes its respective input to generate a modality-specific representation.
The encoders, denoted as $\phi^{a}$ and $\phi^{v}$, extract features from the corresponding modality of $\boldsymbol{x}$.
The outputs are represented as $\boldsymbol{z}^a = \phi^{a}(\theta^{a}, \boldsymbol{x}^{a})$ and $\boldsymbol{z}^{v} = \phi^v(\theta^{v}, \boldsymbol{x}^{v})$, where $\theta^{a}$ and $\theta^{v}$ are the encoder parameters. 
The unimodal outputs are fused \cite{Multisensory, EarlyVsLate} to obtain the multimodal representation.
Without loss of generality, we use concatenation as the fusion method and denote the linear classifier parameters producing the logits by $\boldsymbol{W} \in \mathbb{R}^{C\times (d_{z^a} + d_{z^v})}$ and $\boldsymbol{b} \in \mathbb{R}^{C}$.
The Cross-entropy(CE) loss $\mathcal{L}$ is employed to measure the discrepancy between the model predictions and the targeted label space.

\subsection{Balanced Multimodal Label-space Reshaping}
\subsubsection{Overview of Method}
The overall framework of our method is illustrated in Figure \ref{fig:method}.
We first obtain the unimodal prediction results for each sample and construct the \textbf{Label-space Reshaping Function} $\mathcal{F}$ based on these predictions.
Through $\mathcal{F}$, the original label space $\mathbf{Y}$ is reshaped to produce a balanced and class-relation-aware label space $\mathbf{\widetilde{Y}^u}$, as defined in Equation \ref{eq: y=fy}.
Subsequently, we design a \textbf{Targeted Parameter Optimization} strategy that applies different optimization objectives to distinct parts of the model, enabling simultaneous improvement of unimodal representation learning and multimodal fusion capability.
\begin{equation}
\mathbf{\widetilde{Y}^u} = \mathcal{F}(\mathbf{Y}), \quad u \in \{a, v\}
\label{eq: y=fy}
\end{equation}

\subsubsection{Label-space Reshaping Function}
Considering the variation in modality dependence across samples \cite{resample}, we design a \textbf{sample-level} label-space reshaping strategy based on unimodal predictions.
Specifically, we construct a family of reshaping functions $\mathcal{F} = \{\mathbf{F}_i\}_{i=1}^N$, where $\mathbf{F}_i$ represents the label-space reshaping applied to the $i$-th sample.
Each $\mathbf{F}_i$ is adaptively determined by the unimodal predictions of the sample and the label reshaping threshold $\beta$. 
The reshaping process is formulated as follows:
\begin{equation}
\widetilde{\boldsymbol{y}_i}^u = \mathbf{F}_i(\boldsymbol{y}_i, \boldsymbol{z}_i^a, \boldsymbol{z}_i^v, \beta), \quad u \in \{a, v\}.
\label{eq: yi=fy}
\end{equation}
Each reshaping function $\mathbf{F}_i$ consists of two fundamental components: a \textbf{Reshaping Matrix}, which determines how the label space is reshaped, and a \textbf{Reshaping Intensity}, which controls the degree of reshaping. 
In this section, we elaborate on our design based on these two key aspects.

\textbf{Reshaping Matrix} 
As discussed in Section \ref{sec:introduction}, we argue that modality imbalance arises from differences in the difficulty of mapping between the feature space and the label space.
Therefore, applying appropriate smoothing to the label space offers a potential solution. 
However, naive label smoothing weakens the original supervision signal and may introduce incorrect inter-class relationships Figure \ref{fig: normal_smooth}, resulting in a noticeable performance drop, as shown in Table \ref{tbl:ablation}.
To address this issue, we utilize the predictions from the complementary modality to guide the reshaping of the current label space.

Specifically, we first obtain the unimodal prediction of each sample as OGM\cite{ogm} does and convert them into probability distributions using the \textit{softmax} function, as follows:
\begin{equation}
\boldsymbol{p}_i^u = \text{softmax}(\boldsymbol{W}^{u} \cdot \boldsymbol{z}^{u} +\frac{\boldsymbol{b}}{2}), \quad u \in \{a, v\}.
\label{eq: pred}
\end{equation}
Due to modality imbalance during training, it is necessary to evaluate the degree of discrepancy between modalities and adjust the temperature accordingly to prevent the predictions from becoming overly sharp or overly smooth.
Inspired by the imbalance factor defined in OGM \cite{ogm}, we measure the confidence discrepancy between modalities at the sample level, as shown in Equation \ref{eq: imbalance ratio}:
\begin{equation}
\lambda_i^{a} = \frac{\boldsymbol{y}_i^T \cdot \boldsymbol{p}_i^a}{\boldsymbol{y}_i^T \cdot \boldsymbol{p}_i^v}.
\label{eq: imbalance ratio}
\end{equation}
$\lambda_{i}^{v}$ is accordingly defined as the reciprocal of $\lambda_{i}^{a}$.
Based on the confidence discrepancy, we reshape the unimodal prediction results using a strategy similar to knowledge distillation \cite{DistillSurvey}, thereby injecting cross-modal inter-class information, as shown in Equation \ref{eq: distill}:
\begin{equation}
\begin{aligned}
\boldsymbol{d}_i^{a} &= \text{softmax} (\boldsymbol{W}^v \cdot \boldsymbol{z}^v +\frac{\boldsymbol{b}}{2}, T= \alpha \cdot \lambda_i^v), \\
\boldsymbol{d}_i^{v} &= \text{softmax} (\boldsymbol{W}^a \cdot \boldsymbol{z}^a +\frac{\boldsymbol{b}}{2}, T= \alpha \cdot \lambda_i^a),
\end{aligned}
\label{eq: distill}
\end{equation}
where $\alpha$ is a hyperparameter that regulates the sensitivity of the mapping between confidence $\lambda_i^u$ and temperature $T$, and we decompose $\boldsymbol{W}$ to $[\boldsymbol{W}^a; \boldsymbol{W}^v]$.
Accordingly, the Reshaping Matrix for the label space of each modality can be obtained as follows:
\begin{equation}
\mathbf{D}_i^u = \boldsymbol{d}_i^{u} \cdot \mathbf{1}^T \in \mathbb{R}^{C \times C}, \quad u \in \{a, v\}.
\label{eq: Reshaping Matrices}
\end{equation}

\textbf{Reshaping Intensity}
We reshape the label space to align the mapping difficulty between the feature space and the label space, thereby mitigating modality imbalance.
To ensure sufficient modality alignment while avoiding excessive reshaping that could introduce bias or weaken the supervision signal, we further constrain both the conditions and the intensity of the reshaping process.

Concretely, we define the Reshaping Intensity $\xi$ as follows:
\begin{equation}
\xi _{i}^{u} = 
\begin{cases}
1-\frac{1}{\lambda_{i}^{u}} & 1 < \lambda_{i}^{u} < 1 / \beta \\
0  &  others
\end{cases}, \quad u \in \{a, v\},
\label{eq: xi}
\end{equation}
where $\beta \in [0, 1]$ is a hyperparameter\footnote{Since $\beta$ appears in the denominator, its value cannot be zero. To simulate the case where $1 / \beta$ approaches infinity, we introduce an extremely small positive value as a substitute for $\beta = 0$.}, which controls when reshaping is activated.
The label space of a modality is reshaped only when the confidence ratio is less than $ 1/ \beta$. The key insight is that if the confidence ratio exceeds $1 / \beta$, the complementary modality is poorly learned and should not be used for distillation.
In this case, the original supervision signal is retained for the insufficiently learned modality to ensure adequate learning, as done in previous works \cite{ogm, remix}.

\textbf{Label Space After Reshaping.} Based on the Reshaping Matrix and Reshaping Intensity described above, we perform sample-level label-space reshaping for each modality, as formulated in Equation \ref{eq: reshape_y}:
\begin{equation}
\widetilde{\boldsymbol{y}_i}^u \cdot \mathbf{I} = \mathbf{F}_i(\boldsymbol{y}_i) = (\xi_i^u \cdot \mathbf{D}_i^u + (1-\xi_i^u) \cdot \mathbf{I}) \boldsymbol{y}_i.
\label{eq: reshape_y}
\end{equation}
Here, $\mathbf{I} \in \mathbb{R}^{C \times C}$ denotes the identity matrix, representing the one-hot component preserved during reshaping.
Both the reshaping matrix $\mathbf{D}_i^u$ and the reshaping strength $\xi_i^u$ are functions of the current modality prediction and predefined hyperparameters.
This cross-modal label reshaping strategy not only aligns the mapping difficulty between the feature space and the label space\footnote{$\mathbf{D}_i^u$ and $\mathbf{I} $ are valid probability distributions(non-negative, sum to one), which guarantee the reshaped label space to be valid.} across modalities but also captures inter-class relationships, thereby enhancing modality interaction (as demonstrated in Section \ref{sec: cra}).

\subsubsection{Targeted Parameter Optimization}
During training, we dynamically reshape the label space for each modality at the sample level.
Consequently, applying a standard joint optimization strategy under these conditions can lead to fluctuations in the optimization objectives, which may hinder model convergence and effective learning.
Inspired by AMSS \cite{amss} and DGL \cite{dgl}, we group the model parameters and assign different optimization objectives to specific parameter sets, ensuring more stable and effective training.

In multimodal learning, modality imbalance often leads existing approaches to assign each modality an independent classification head and compute a separate unimodal loss \cite{makes}.
These losses are then combined through weighted fusion into a joint loss $\mathcal{L}$, which is used to optimize all model parameters simultaneously, as follows:
\begin{equation}
\mathcal{L} = \mathcal{L}_{CE}^{0} + \sum_{u \in \{a, v\}} w^{u}\mathcal{L}_{CE}^u,
\label{eq: L_initial}
\end{equation}
where $\mathcal{L}_{CE}^0$ denotes the multimodal loss, while $w^{k}$ and $\mathcal{L}_{CE}^k$ represent the weight and the unimodal loss for the $k$-th modality, respectively.
In standard learning, the final optimization target always corresponds to the ground-truth one-hot labels, making global optimization straightforward. 
However, since our method operates from thon the label side, the target labels across modalities and samples can be inconsistent and dynamically change training, as visualized in Section \ref{sec: vis}. 
Applying a unified global optimization under these conditions can lead to model instability or even prevent convergence.
To address this, we group the model parameters and update each targeted group with a specific objective.

Specifically, we define two types of learning objectives: $\mathcal{L}^u$ computes the loss between the reshaped label space and the output of each modality, while $\mathcal{L}^0$ measures the loss between the final multimodal output and the original one-hot labels, as formulated below:
\begin{equation}
\mathcal{L}^u = -\sum_i^N \widetilde{\boldsymbol{y}_i}^u \log (\boldsymbol{p}_i^u), 
\quad
\mathcal{L}^0 = -\sum_i^N \boldsymbol{y}_i \log (\boldsymbol{p}_i), 
\label{eq: Loss}
\end{equation}
where $\boldsymbol{p}_i$ denotes the multimodal prediction. 
During training, the loss computed from the reshaped label space is used exclusively to optimize the corresponding modality encoder, while the multimodal loss is applied to update the parameters of the decision layer, as shown in Equation \ref{eq: update}:
\begin{equation}
\theta^{u,*}  = \arg \min_{\theta^u} \mathcal{L}^u, 
\quad
\Theta^*  = \arg \min_{\Theta} \mathcal{L}^0.
\label{eq: update}
\end{equation}
Here, $\Theta = \{\boldsymbol{W}, \boldsymbol{b}\}$ denotes the set of decision layer parameters. 
This approach anchors the primary learning objectives for each part of the model, reducing potential biases that could arise from simply summing all losses. 

\subsubsection{Summary and Pseudo-code}
In summary, the pseudo-code of BMLR is presented in Algorithm \ref{alg: bmlr}. 
Our method simultaneously harmonizes the learning dynamics across modalities and enhances the unimodal performance by generating labels enriched with richer informational and relational cues.

\begin{algorithm}[t]
\caption{Method of BMLR}
\label{alg: bmlr}
\begin{algorithmic}[0]
    \STATE {\bfseries Input:} data $\mathcal{D}=\{(\boldsymbol{x}_{i}^{a}, \boldsymbol{x}_{i}^{v}), \boldsymbol{y}_{i}\}_{i=1,2,...,N}$, modality set $U = \{a, v\}$, training epoch $E$, initialized model parameters $\theta^a$, $\theta^v$, $\Theta$, learning rate $\eta$, hyperparameters $\alpha$ ,$\beta$
    \FOR{$e = 0$, $\cdots$, $E-1$}
    \FOR{each sample $\boldsymbol{x}$ in $\mathcal{D}$}
    \STATE Get predict results $\boldsymbol{p}_i^u$ with Equation \ref{eq: pred};
    \STATE Calculate confidence discrepancy $\lambda_i^u$ with Equation \ref{eq: imbalance ratio};
    \STATE Build the Reshaping Matrix $\mathbf{D}^u_i$ with Equation \ref{eq: Reshaping Matrices};
    \STATE Calculate Reshaping Intensity $\xi^u_i$ with Equation \ref{eq: xi};
    \STATE Reshaping the label space with $\mathbf{D}^u_i$ and $\xi^u_i$;
    \ENDFOR
    \FOR{$u$ in $U$}
        \STATE Calculate the unimodal loss $\mathcal{L}^u$;
        \STATE $\theta^{u,*} \gets \theta^u - \eta \nabla_{\theta^u}\mathcal{L}^u $;
    \ENDFOR
    \STATE Calculate the multimodal loss $\mathcal{L}^0$;
    \STATE $\Theta^{*} \gets \Theta - \eta \nabla_{\Theta}\mathcal{L}^0 $;
\ENDFOR
\end{algorithmic}
\end{algorithm}

\newcommand{\xsmall}[1]{{\fontsize{8pt}{8pt}\selectfont #1}}
\renewcommand{\arraystretch}{1.2}
\begin{table*}[t]
\centering
\caption{\textbf{Comparison with Balanced Multimodal Learning Methods.} 
\textcolor{red}{\textbf{Red}} represents the best accuracy in existing methods, while \textcolor{blue}{\textbf{Blue}} represents the second-best performance.}
\label{tbl:method_comparison}
\begin{tabular}{
@{}
>{\centering\arraybackslash}p{4.5cm}|
>{\centering\arraybackslash}p{3cm}|
>{\centering\arraybackslash}p{3cm}|
>{\centering\arraybackslash}p{3cm}
@{}}
\toprule
\textbf{Method}  & \textbf{CREMAD} & \textbf{Kinetic-Sounds} & \textbf{AVE} \\ 
\midrule
Baseline &  64.35\% & 63.59\% & 67.19\% \\
Uniform-Baseline & 70.88\% & 68.00\% & 70.52\% \\
\midrule
OGM-GE \xsmall{(CVPR 2022)} & 65.19\% & 66.52\% & 62.76\% \\ 
AGM \xsmall{(ICCV 2023)} & 72.44\% & 63.78\% & 66.41\% \\ 
CML \xsmall{(ICML 2023)} & 67.61\% & 68.13\% & 69.27\% \\ 
MBSD \xsmall{(WSDM 2023)} &  72.87\% & 65.16\% & 70.05\% \\ 
LFM \xsmall{(NuerIPS 2024)} & 63.78\% & 64.26\% & 65.89\% \\
Diag\&Re \xsmall{(ECCV 2024)} & 66.62\% & 63.75\% & 63.28\% \\
ReconBoost \xsmall{(ICML 2024)} & 68.61\% & 67.89\% & \textcolor{blue}{\textbf{74.22\%}} \\
MMPareto \xsmall{(ICML 2024)} &  \textcolor{blue}{\textbf{74.57\%}} & 68.75\% & 71.09\% \\
OPM \xsmall{(TPAMI 2025)} & 65.48\% & 65.82\% & 66.41\% \\
Remix \xsmall{(ICML 2025)} & 72.16\% & 68.47\% & 68.23\% \\ 
DGL \xsmall{(ICCV 2025)} & 71.30\% & \textcolor{blue}{\textbf{68.86\%}} & 68.75\% \\
\midrule
\textbf{BMLR} \xsmall{(ours)} & \textcolor{red}{\textbf{79.69\%}} & \textcolor{red}{\textbf{71.76\%}} & \textcolor{red}{\textbf{76.04\%}} \\
\bottomrule
\end{tabular}
\end{table*}

\section{Experiments}
\subsection{Experimental Details}
\subsubsection{Datasets}
The main experiments in this work are conducted on three multimodal video datasets, following the prior studies \cite{ogm, bml}.
\textbf{CREMAD} \cite{cremad} is an audiovisual dataset for emotion recognition, consisting of 7,440 video clips.
It includes six common emotions, and the dataset is randomly split into a training and validation set of 6,696 samples and a test set of 744 samples, with an approximate 9:1 ratio.
\textbf{Kinetic-Sounds (K-Sounds)} \cite{ks} is derived from the Kinetics dataset \cite{kinetic}, which contains 400 action classes based on YouTube videos.
Kinetic-Sounds consists of 31 action categories selected for their potential to be represented both visually and aurally, such as playing various instruments.
\textbf{AVE} \cite{ave} is a subset of the AudioSet dataset designed for audio-visual event localization. It contains 28 event categories, totaling 4,143 10-second videos. 
The training, validation, and test sets include 3,339, 402, and 402 samples, respectively.

To further demonstrate the effectiveness of BMLR under settings with more than two modalities, we conduct experiments on two tri-modal sentiment analysis datasets.
\textbf{CMU-MOSI (MOSI)} \cite{mosi} is an early benchmark dataset for multimodal sentiment analysis, consisting of 2,199 clips from 93 YouTube opinion videos with aligned text, audio, and visual modalities. Each utterance is annotated with a sentiment intensity score from -3 to +3 for classification.
\textbf{CMU-MOSEI (MOSEI)} \cite{mosei} is a large-scale extension of MOSI, which contains 23,453 videos with synchronized text, audio, and visual modalities, annotated with the same labels (-3 to +3).

\subsubsection{Experimental Settings}
For bi-modal datasets (CREMAD, K-Sounds, AVE), we adopt ResNet-18 \cite{resnet} trained from scratch to extract features for all modalities. 
For the video modality, frames are sampled at 1 fps and uniformly selected as inputs.
For the audio modality, spectrograms are generated using Librosa \cite{librosa} and serve as the input. 
For tri-modal datasets (MOSI, MOSEI), we adopt the Transformer-based \cite{transformer} model for feature processing and fusion. 
During training, the Adam \cite{adam} optimizer is used for parameter optimization, with $\beta = (0.9, 0.999)$ and a learning rate set to $5 \times 10^{-5}$. 
All reported results are averages from three random seeds, with all models trained on one NVIDIA RTX 3090 GPUs using a batch size of 64 for 70 epochs.

\begin{table}[t]
\centering
\caption{\textbf{Combination and comparison with conventional fusion methods.}
$\dagger$ indicates that our BMLR method is applied. \textbf{Bold} indicates that our method brings improvement.}
\label{tbl:fusion}
\begin{tabular}{
@{}>{\centering\arraybackslash}p{1.5cm}|
>{\centering\arraybackslash}p{1.8cm}|
>{\centering\arraybackslash}p{1.8cm}|
>{\centering\arraybackslash}p{1.7cm}
@{}}
\toprule
\textbf{Method} & \textbf{CREMAD} & \textbf{K-Sounds} & \textbf{AVE} \\ 
\midrule
Concat   & 70.88\%   & 68.00\% & 70.52\% \\
Concat$^\dagger$   & \textbf{79.69\%}   & \textbf{71.76\%} & \textbf{76.04\%} \\
\midrule
Sum       & 70.88\%   & 68.75\%  & 66.67\% \\
Sum$^\dagger$       & \textbf{79.26\% }  & \textbf{71.95\%}  & \textbf{75.52\%} \\
\midrule
Film            & 63.34\%   & 64.65\%  & 68.23\% \\
Film$^\dagger$            & \textbf{68.47\% }  & \textbf{67.97\%}  & \textbf{73.44\%} \\ 
\midrule
Bi-Gated           & 63.92\%   & 63.36\%  & 66.93\% \\
Bi-Gated$^\dagger$           & \textbf{65.63\%}   & \textbf{66.24\%}  & \textbf{73.70\%} \\
\bottomrule
\end{tabular}
\end{table}

\subsection{Comparison with Existing BML Methods}
Our BMLR method is designed to alleviate modality imbalance in multimodal learning from the label perspective. 
To validate its effectiveness and superiority, we compare it with existing balanced learning methods, as shown in Table \ref{tbl:method_comparison}. 
Specifically, the methods include \textbf{OGM-GE} \cite{ogm}, \textbf{AGM} \cite{agm}, \textbf{CML} \cite{cml}, \textbf{MBSD} \cite{mbsd}, \textbf{LFM} \cite{lfm}, \textbf{Diag\&Re} \cite{Diagnosing}, \textbf{ReconBoost} \cite{reconboost}, \textbf{MMPareto} \cite{mmpareto}, \textbf{OPM} \cite{opm}, \textbf{Remix} \cite{remix}, and \textbf{DGL} \cite{dgl}.
\textbf{Baseline} denotes training the model using only the multimodal loss, whereas \textbf{Uniform-Baseline} optimizes with a weighted combination of unimodal and multimodal losses \cite{makes}.
The results demonstrate that BMLR achieves remarkable performance improvements across all datasets. 

\subsection{Combination with Different Fusion Methods}
We compare our approach with several representative multimodal fusion strategies commonly used in deep learning frameworks, including \textbf{Concatenation (Concat)} \cite{concat}, \textbf{Summation (Sum)}, \textbf{FiLM} \cite{film}, and \textbf{Bi-Gated} \cite{gated}. The results are shown in Table \ref{tbl:fusion}.
Following prior works, we adopt the combination of BMLR and concatenation as a representative method.
The results clearly demonstrate that our approach achieves substantial performance improvements across datasets, with gains of \textbf{8.81\%} on CREMAD, \textbf{3.76\%} on Kinetic-Sounds, and \textbf{5.52\%} on AVE.
Moreover, our method consistently yields significant performance gains when combined with various fusion strategies, further demonstrating its general applicability.

\begin{table*}[t]
\centering
\caption{Comparison of multimodal accuracy, unimodal accuracy, and A/V Ratio before and after applying BMLR. The best unimodal performance is \underline{underlined}. For the Ratio metric, a downward arrow (\textcolor{red}{\textbf{$\downarrow$}}) indicates that values closer to 1.}
\label{tab:unimodal}
\setlength{\tabcolsep}{12pt}
\renewcommand{\arraystretch}{1.2}
\begin{tabular}{ccccc||cccc}
\toprule
& \multicolumn{4}{c}{Baseline} & \multicolumn{4}{c}{BMLR (Ours)} \\
\cmidrule(lr){2-5} \cmidrule(lr){6-9}
\textbf{Dataset} & Multimodal & Audio & Visual & A/V Ratio & Multimodal & Audio & Visual & A/V Ratio\\
\midrule
\textbf{CREMAD}     & 70.88\% & 54.12\% & 31.25\% & \textbf{1.73} & 79.69\% & \underline{62.22\%} & \underline{66.34\%} & \textbf{0.94} \textcolor{red}{\textbf{$\downarrow$}}\\
\textbf{K-Sounds}   & 68.99\% & 49.41\% & 30.63\% & \textbf{1.61} & 71.76\% & \underline{52.66\%} & \underline{54.49\%} & \textbf{0.97} \textcolor{red}{\textbf{$\downarrow$}}\\
\textbf{AVE}        & 70.52\% & 58.33\% & 24.47\% & \textbf{2.38} & 76.04\% & \underline{68.23\%} & \underline{38.80\%} & \textbf{1.76} \textcolor{red}{\textbf{$\downarrow$}}\\
\bottomrule
\end{tabular}
\end{table*}

\subsection{Analysis of the BMLR Method}

\subsubsection{Analysis of the Alleviation of Modality Imbalance and Promotion of Cross-modal Interaction}\label{sec: uni} 
To verify the effectiveness of our method in mitigating modality imbalance and promoting cross-modal interaction, we further report the multimodal accuracy, unimodal accuracy (Audio / Visual), and modality performance ratio (A/V Ratio) before and after applying BMLR. 
Specifically, the A/V Ratio denotes the ratio between audio and visual accuracies, which is used to measure the performance gap between modalities. A value closer to 1 indicates a more balanced contribution of different modalities.

As shown in Table \ref{tab:unimodal}, after introducing BMLR, the multimodal performance is consistently improved on all three datasets. Meanwhile, the A/V Ratio is significantly reduced and becomes closer to 1, indicating that the model's over-reliance on the dominant modality is effectively alleviated and \textbf{the modality learning becomes more balanced}.
In addition, the performance of both unimodal branches is improved, with larger gains observed in the weak modality. 
At the same time, the strong modality is not suppressed but further improved.
This indicates that BMLR does not achieve balance by sacrificing the strong modality, but instead \textbf{promotes cross-modal interaction and collaborative optimization}.

Overall, our method not only improves the overall multimodal recognition performance but also effectively narrows the performance gap between modalities and strengthens cross-modal interaction, leading to a more balanced and efficient multimodal learning process.

\subsubsection{Analysis of the Balanced and Class-Relationship-Aware Label Space}\label{sec: vis}
Our BMLR explicitly reshapes the supervision signals for each modality, leading to a more balanced label distribution while preserving inter-class relationships.

\begin{figure}[t]
\centering
\subfloat[Count of Reshaped Samples and Visualization of Label Space.]{\includegraphics[width=0.47\textwidth]{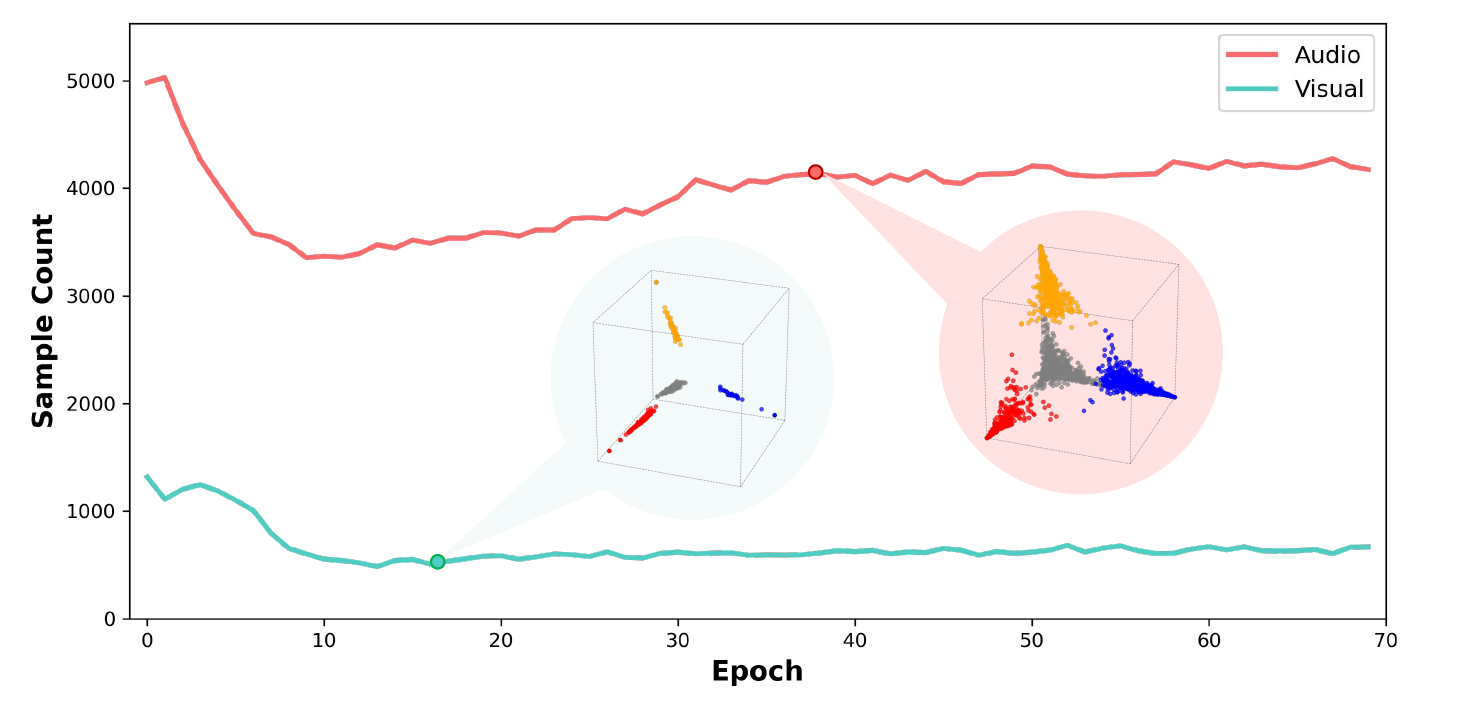}\label{fig: reshape_count}}
\hfil
\subfloat[Uniform]{\includegraphics[width=0.16\textwidth]{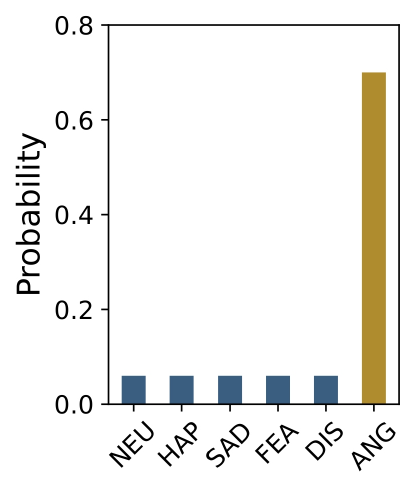}\label{fig: normal_smooth}}
\hfil
\subfloat[BMLR]{\includegraphics[width=0.16\textwidth]{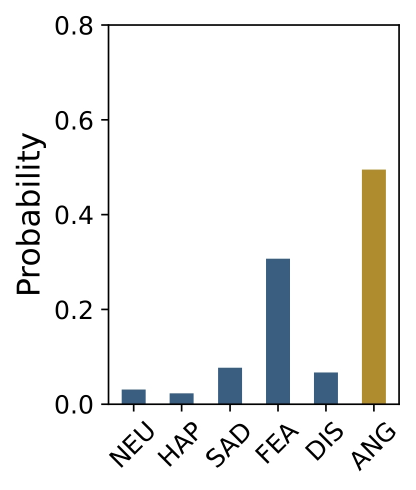}\label{fig: bmlr_smooth}}
\hfil
\subfloat[Sample]{\includegraphics[width=0.13\textwidth]{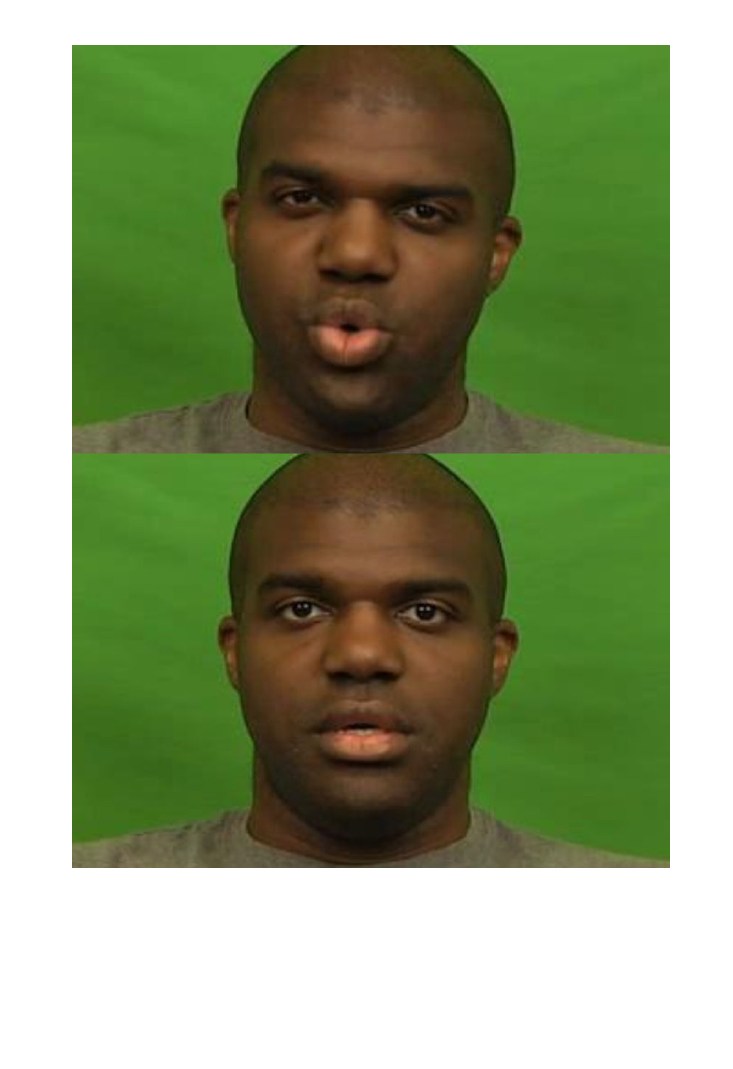}\label{fig: sample}}
\caption{Visualization of the number of samples used for label reshaping and the reshaped label space results.}
\label{fig:smooth_label}
\vskip -0.2in
\end{figure}

\textbf{Balanced Label Space.} 
As shown in Figure \ref{fig: reshape_count}, we analyze the number of samples undergoing label-space reshaping during training and visualize the corresponding label distributions with t-SNE \cite{t-sne}.
For clarity, three classes are selected, where each axis represents the target probability of one class. 
We observe that the audio modality, which typically learns faster and tends to memorize data, receives a broader and more diverse set of soft labels.
This slows down its learning process and helps it capture richer inter-class relationships. 
In contrast, the visual modality learns from fewer and sharper labels, ensuring strong supervision while applying moderate smoothing.
Compared with the original one-hot space, the reshaped label space is substantially more balanced.

\begin{figure}[t]
\centering
\subfloat[Disappointed (V)]{\includegraphics[width=0.23\textwidth]{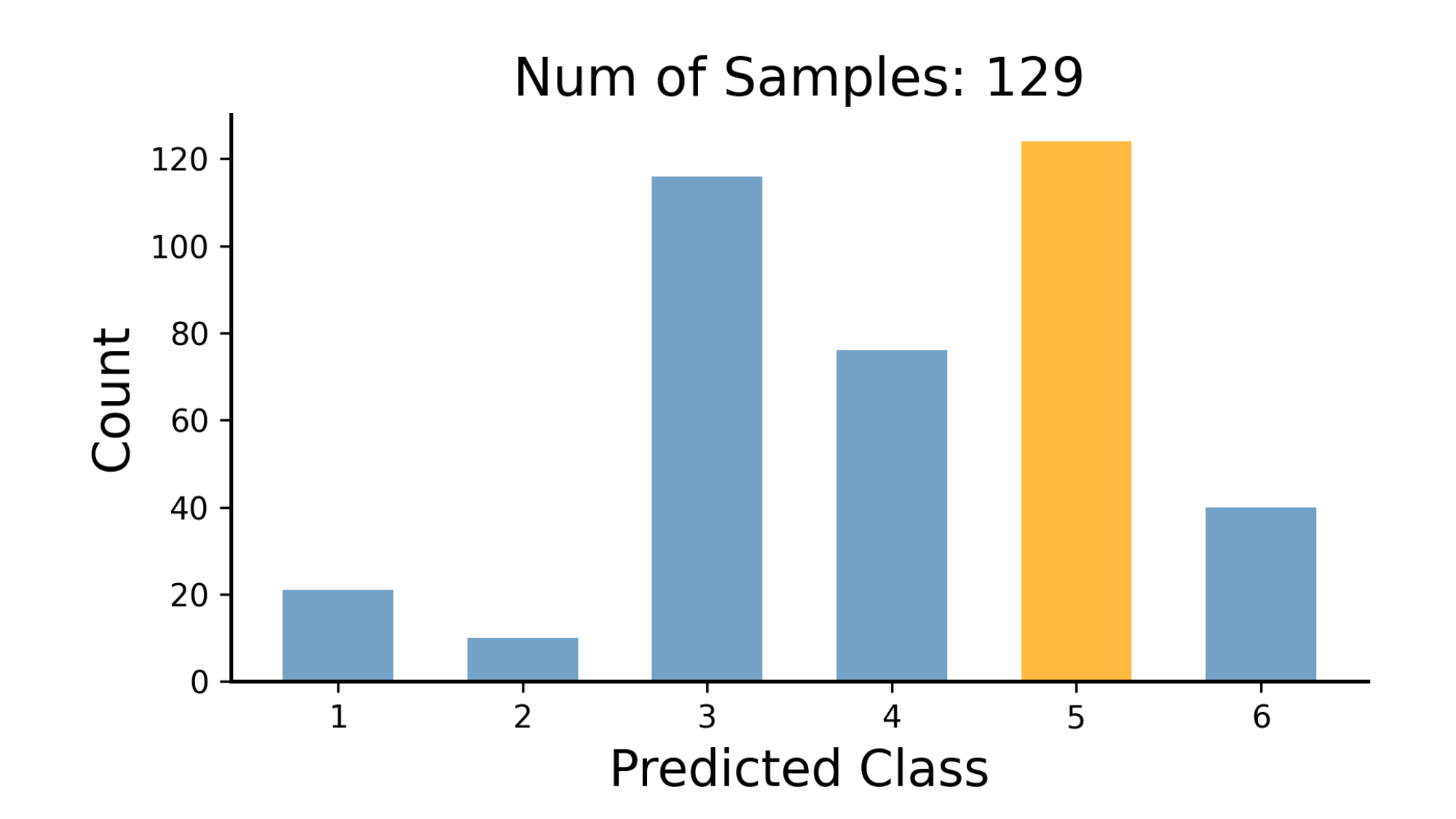}\label{fig:lg_cd_dis}}
\hfil
\subfloat[Happy (V)]{\includegraphics[width=0.23\textwidth]{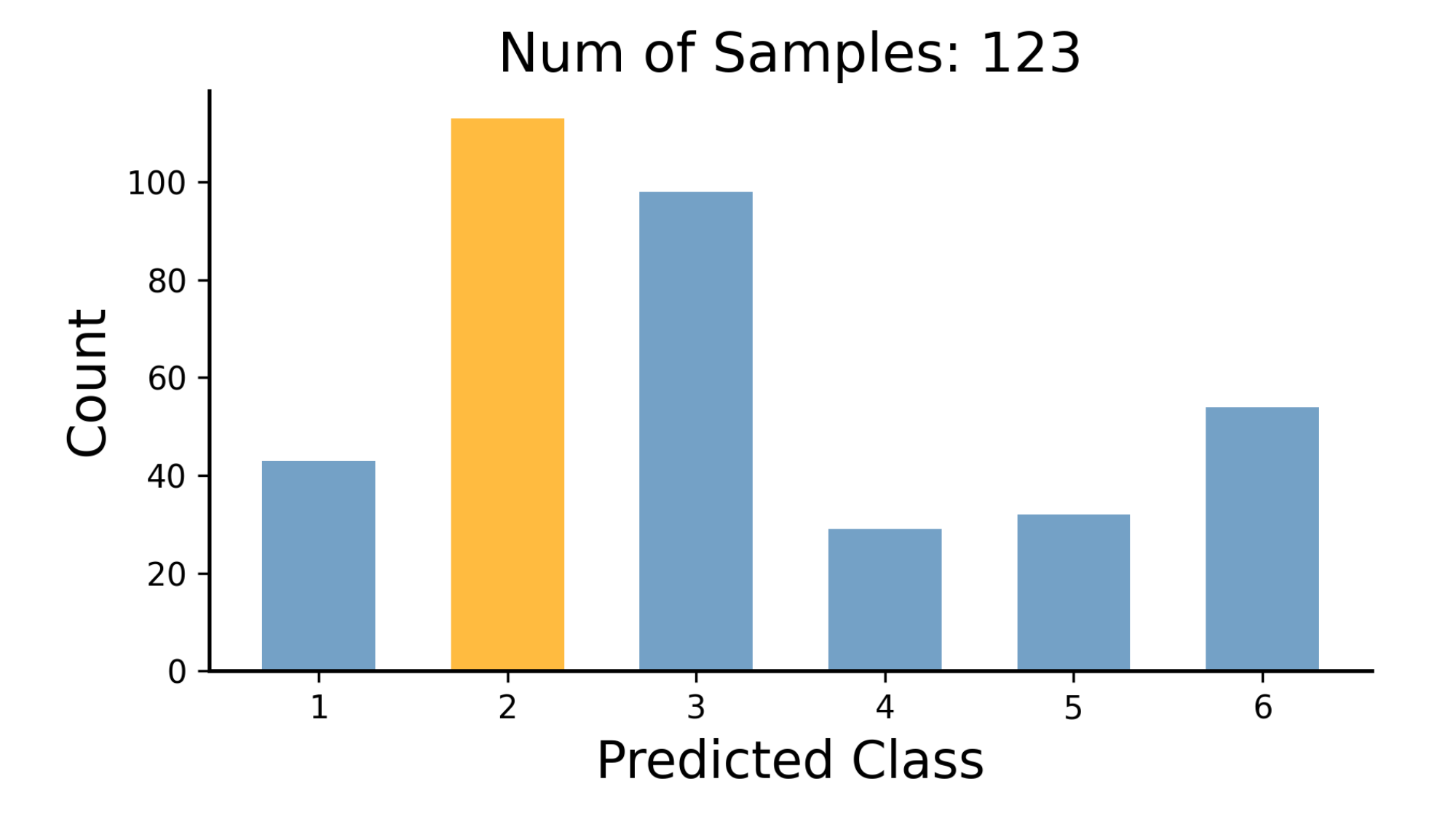}\label{fig:lg_cd_hap}}
\hfil
\subfloat[Playing Clarinet (A)]{\includegraphics[width=0.23\textwidth]{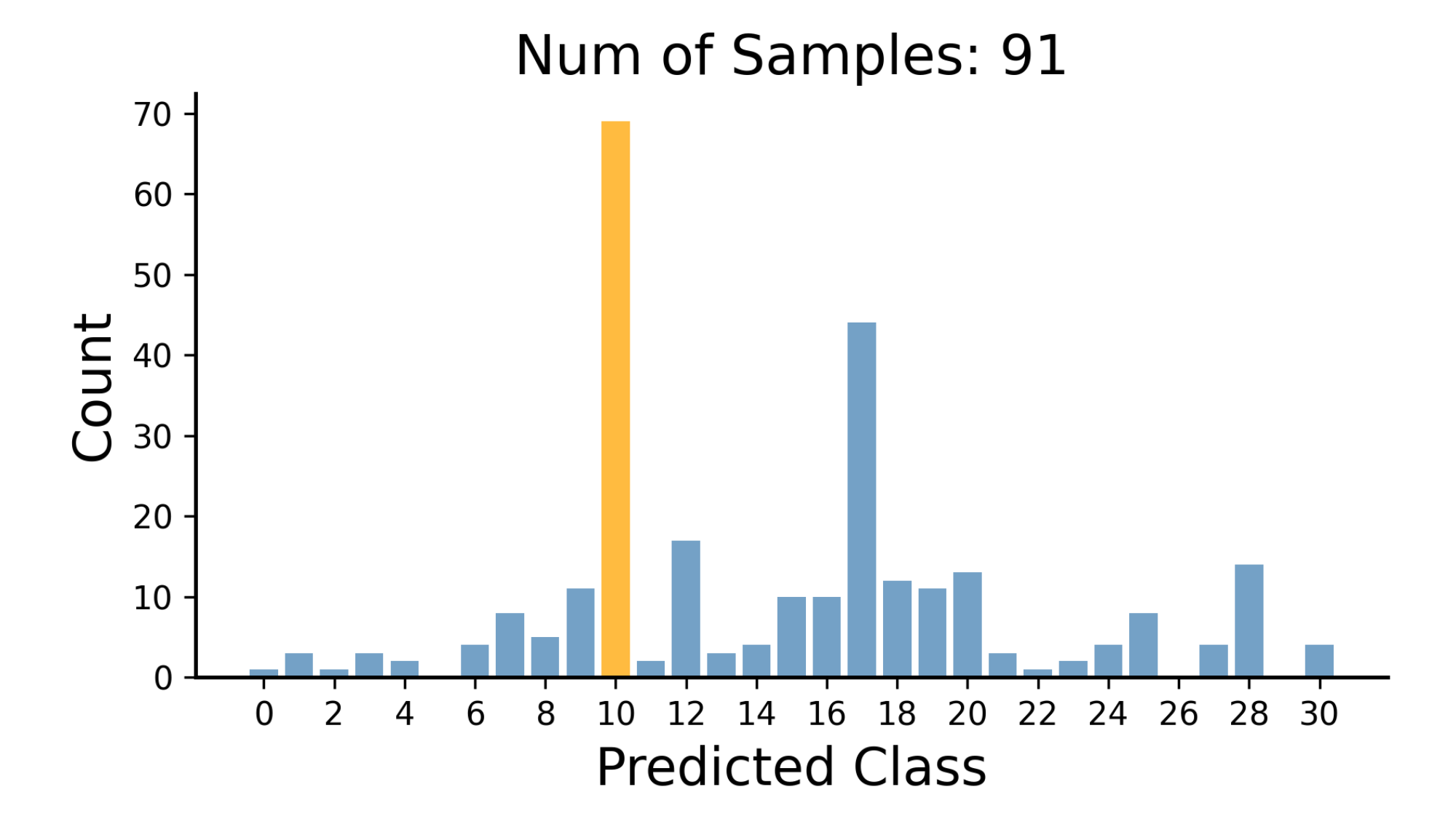}\label{fig:lg_ks_pc}}
\hfil
\subfloat[Playing Saxophone (A)]{\includegraphics[width=0.23\textwidth]{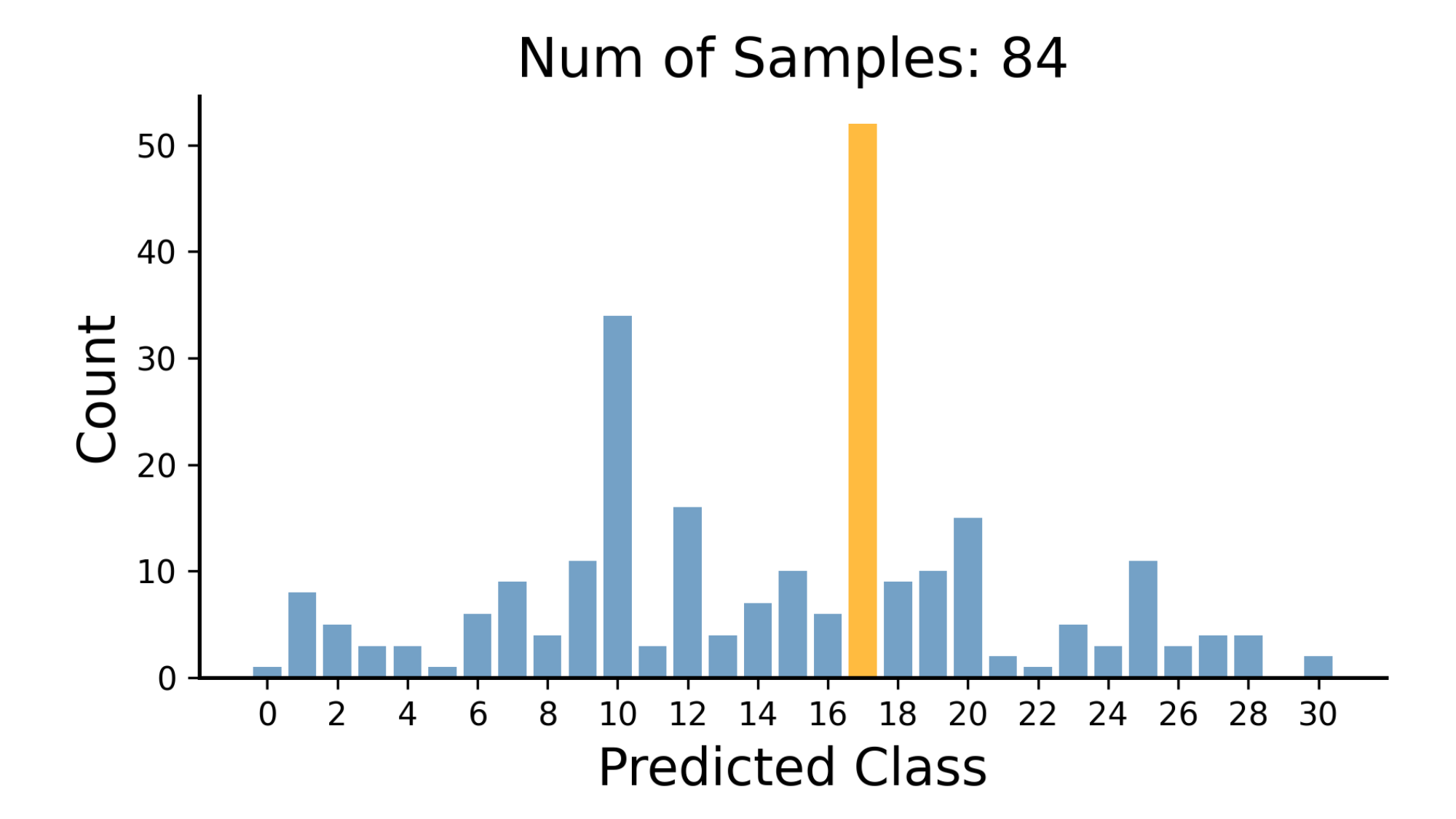}\label{fig:lg_ks_ps}}
\caption{\textbf{Top-3 predicted class frequency distributions for a given class.} Top: visual modality on CREMAD. Bottom: audio modality on KS. \textcolor{Dandelion}{\textbf{Yellow bars}} indicate the \textcolor{Dandelion}{\textbf{ground-truth}} class.}
\label{fig:valuation}
\vskip -0.1in
\end{figure}

\textbf{Class-Relation-Aware Label Space.}
During label-space reshaping, our goal is to construct a label space that accurately reflects inter-class relationships.
In Figure \ref{fig: sample}, samples from the classes \textit{ANG} (top) and \textit{FEA} (bottom) exhibit higher similarity.
When using the Uniform Reshaping  (Uni-Reshaping) method mentioned in LFM \cite{lfm}, which distributes a small equal probability across all non-target classes. As shown in Figure \ref{fig: normal_smooth}, the reshaped labels fail to capture these inter-class relationships and sample-specific differences.
In contrast, our BMLR method generates labels that faithfully preserve the inter-class relationship as shown in Figure \ref{fig: bmlr_smooth}.

\begin{figure*}[!t]
\centering
\subfloat[CREMAD]{\includegraphics[width=0.31\textwidth]{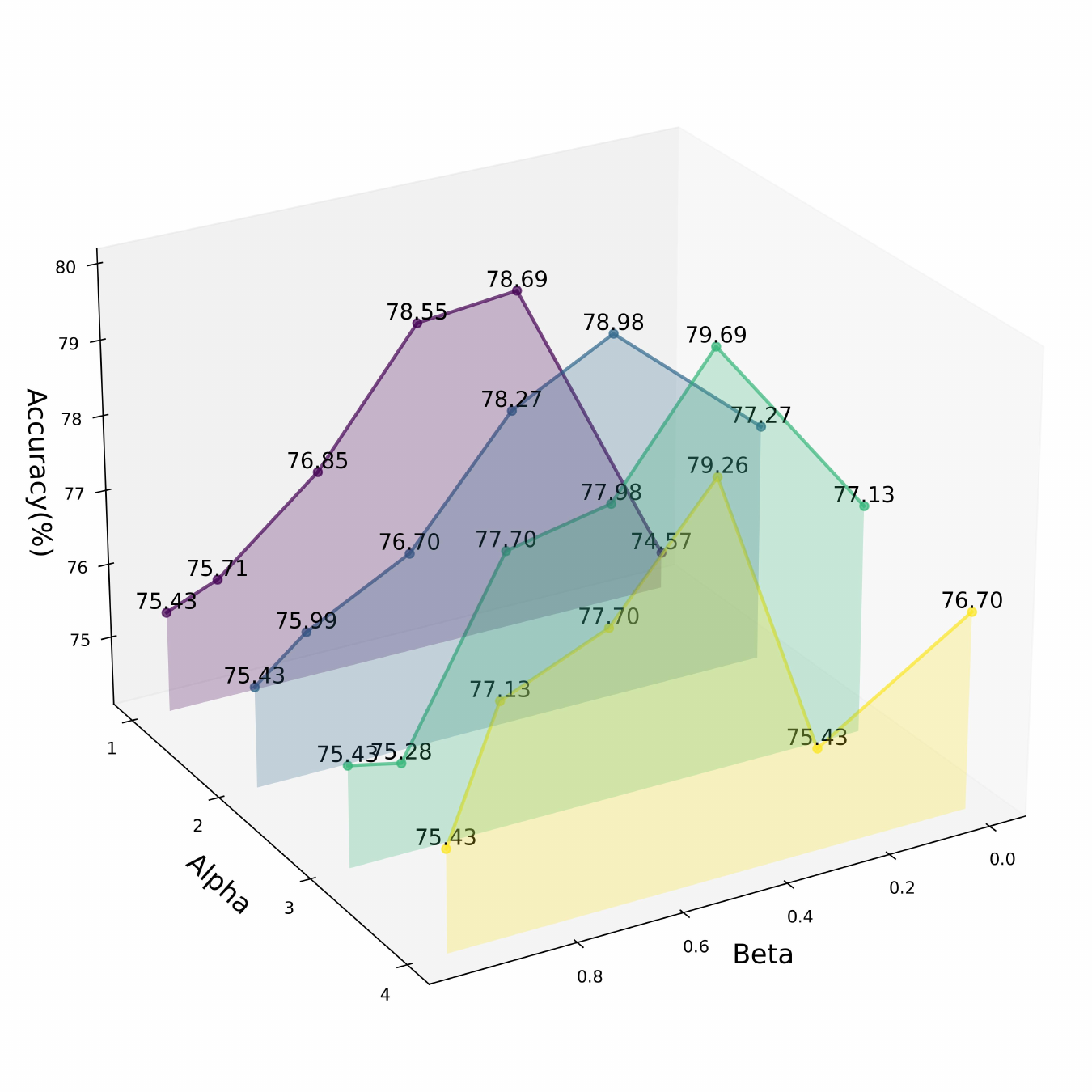}\label{fig:ab_cd}}
\hfil
\subfloat[Kinetic-Sounds]{\includegraphics[width=0.31\textwidth]{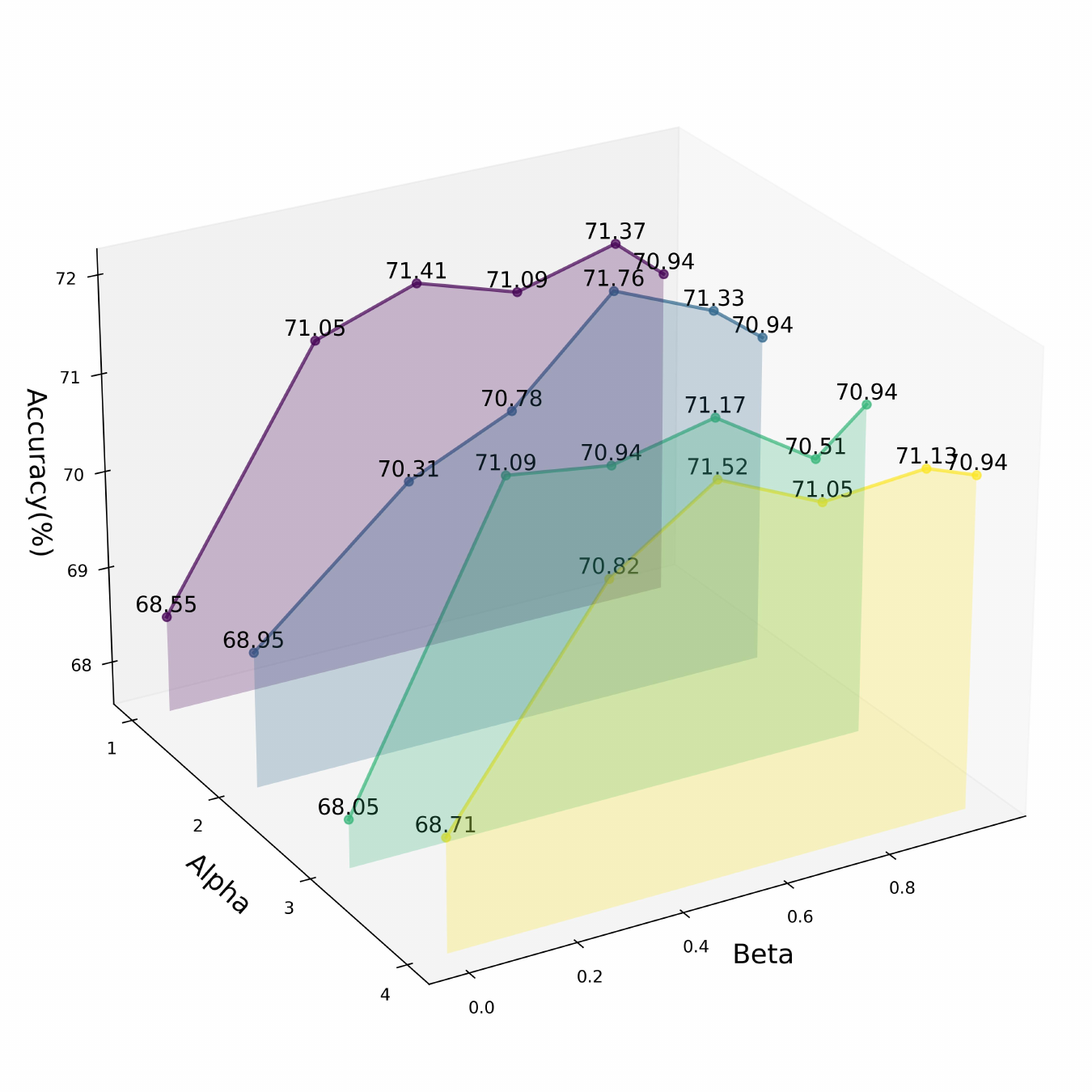}\label{fig:ab_ks}}
\hfil
\subfloat[AVE]{\includegraphics[width=0.31\textwidth]{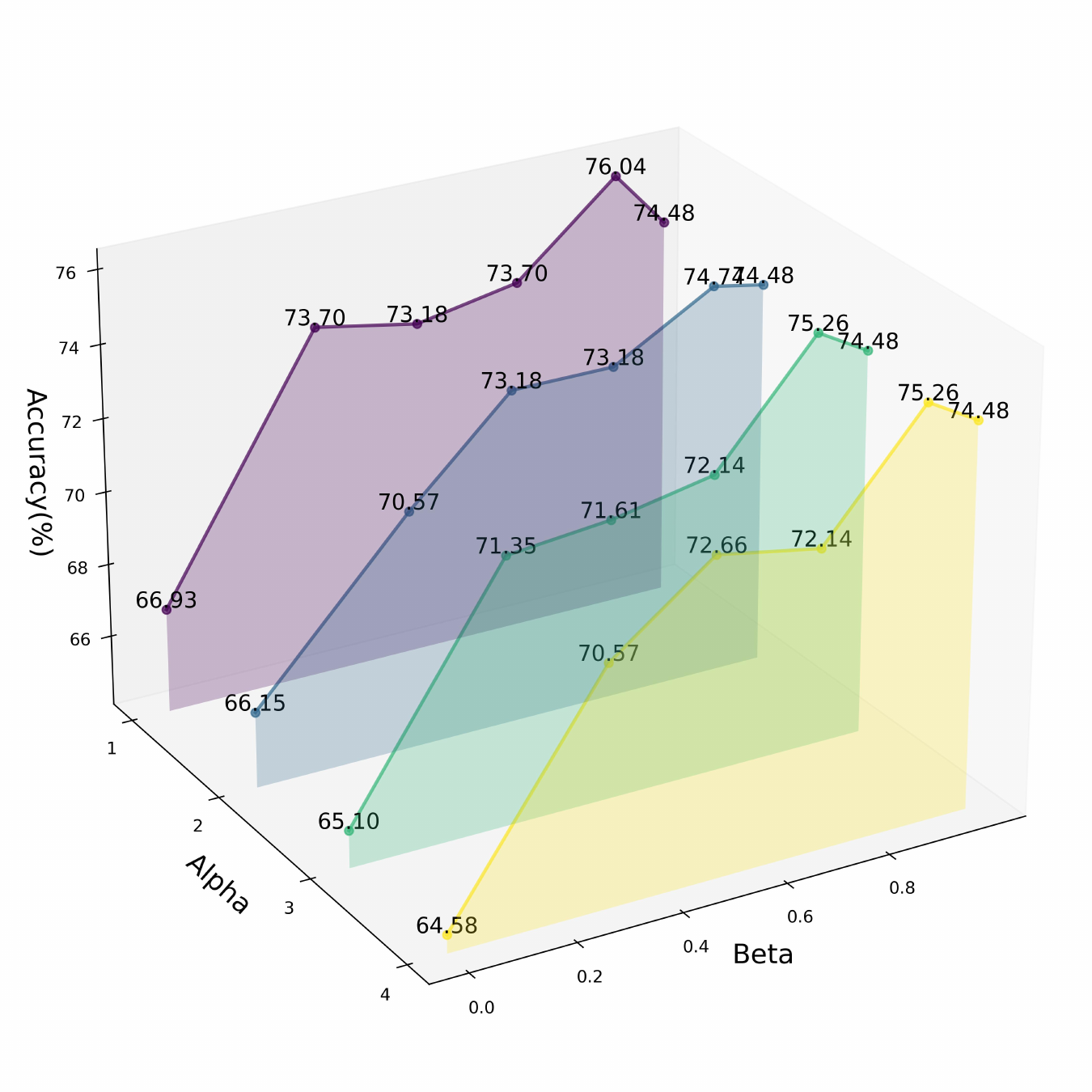}\label{fig:ab_ave}}
\caption{\textbf{Performance of the models with different $\alpha$ and $\beta$ values in BMLR} on different datasets.}
\label{fig:hyperparameters}
\vskip -0.1in
\end{figure*}

\subsubsection{Analysis of Inter-Class Relationships}\label{sec: cra}
We construct a reshaped label space inspired by cross-modal knowledge distillation to enable each modality to better capture inter-class relationships.
To validate the effectiveness of this approach, we conduct an in-depth analysis of the unimodal predictions of models trained with BMLR. 
Specifically, we record the top-3 predicted classes for all samples in each modality and summarize them according to their ground-truth classes.
As shown in Figure \ref{fig:lg_cd_dis} and Figure \ref{fig:lg_cd_hap}, we present representative visual modality results for the \textit{Disappointed} and \textit{Happy} categories from the CREMAD dataset. 
Since these two classes exhibit clear visual differences, their predictions are almost mutually exclusive.  
In contrast, on the Kinetic-Sounds dataset, we display the audio modality results for \textit{Playing Clarinet} and \textit{Playing Saxophone} in Figure \ref{fig:lg_ks_pc} and Figure \ref{fig:lg_ks_ps}, which show strong acoustic correlations and frequent co-occurrence during classification.
This demonstrates that our BMLR method effectively encourages the model to capture inter-class relationships, thereby enhancing performance.

\subsection{Ablation Study}

\subsubsection{Analysis of Hyperparameters in BMLR.} 
The BMLR method introduces two hyperparameters, $\alpha$ and $\beta$.
The parameter $\alpha$ controls the mapping between the confidence discrepancy factor and the temperature, while $\beta$ serves as a threshold for deciding whether label-space reshaping is applied. 
To assess their influence, we conduct a two-parameter sensitivity analysis, as shown in Figure \ref{fig:hyperparameters}.
For each $\alpha$, we vary $\beta$ and visualize the results as multi-layer surface plots.
The overall trend shows that performance initially increases and then decreases.
The parameter $\beta$ governs the timing of label-space reshaping.
When $\beta$ is too small, excessive distillation occurs, leading to performance degradation.
When $\beta$ is too large, label-space reshaping is rarely triggered because $1/\beta$ is close to 1, making reshaping difficult or even impossible, which leads to a performance drop.
In contrast, the effect of $\alpha$ is relatively minor when $\beta$ is fixed, as it only adjusts the mapping between the distillation temperature and the balance factor.
Therefore, once an appropriate $\beta$ is chosen, only a small-scale search over $\alpha$ is required, demonstrating that BMLR does not introduce excessive hyperparameter tuning overhead.

\begin{table}[t]
\centering
\caption{\textbf{Performance of the models trained with different reshaping methods.}  \textcolor{blue}{$\blacktriangle$} and \textcolor{red}{$\blacktriangledown$} indicate performance improvement and degradation compared to the baseline.}
\label{tbl:ablation}
\renewcommand{\arraystretch}{1.2}
\begin{tabular}{
@{}>{\centering\arraybackslash}p{2.6cm}|
>{\centering\arraybackslash}p{2.4cm}|
>{\centering\arraybackslash}p{2.4cm}
@{}}
\toprule
\textbf{Method} & \textbf{CREMAD} & \textbf{K-Sounds} \\ 
\midrule
Baseline      & 70.88\%  & 68.00\% \\
Uni-Reshaping          & 58.95\%  \textcolor{red}{\scriptsize$\blacktriangledown$11.93\%}  & 60.39\%  \textcolor{red}{\scriptsize$\blacktriangledown$7.61\%} \\
BMLR & 79.69\% \textcolor{blue}{\scriptsize$\blacktriangle$8.81\%} & 71.76\% \textcolor{blue}{\scriptsize$\blacktriangle$3.76\%}\\
\bottomrule
\end{tabular}
\vskip -0.1in
\end{table}

\subsubsection{Comparison with Uniform Reshaping Method.} 
\begin{figure}[t]
\centering
\subfloat[Train w/o BMLR]{\includegraphics[width=0.235\textwidth]{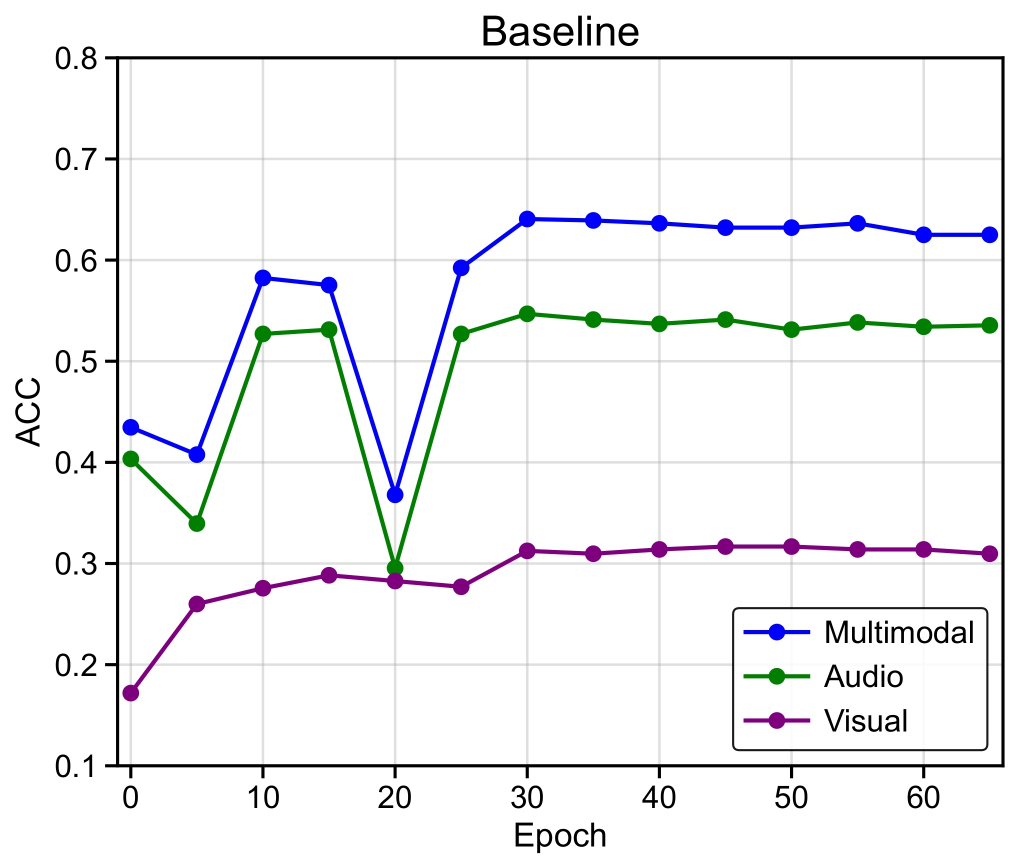}}\label{fig:l_base}
\hfil
\subfloat[Train w/ BMLR]{\includegraphics[width=0.235\textwidth]{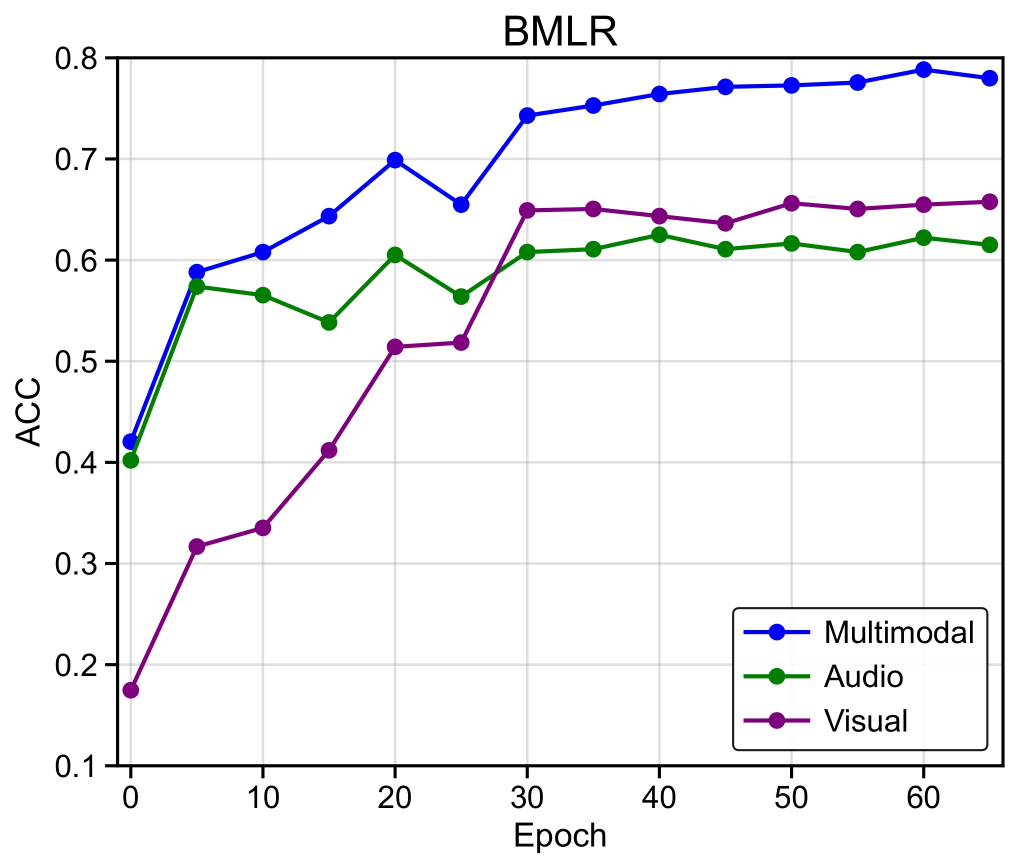}\label{fig:l_bmlr}}
\caption{Learning process before and after applying the BMLR method.}
\label{fig:learning}
\vskip -0.1in
\end{figure}

To verify the effectiveness and advantages of the label-space reshaping strategy in our BMLR method, we first implement the Uniform Reshaping method.
As mentioned in Section \ref{sec: vis}, this simple smoothing weakens the supervision signal and fails to account for sample-specific inter-class relationships, leading to a significant performance drop, as illustrated in Table \ref{tbl:ablation}.
In contrast, our BMLR method employs a more refined label-space reshaping design that incorporates both sample-specific differences and inter-class relationships.
As shown in Figure \ref{fig:learning}, comparing results before and after applying BMLR reveals that our approach not only \textbf{aligns learning difficulty across modalities to mitigate modality laziness} but \textbf{enhances unimodal performance by promoting inter-modal interaction and capturing inter-class relationships}.

\begin{table}[t]
\centering
\caption{Performance of the proposed BMLR and its variants.}
\renewcommand{\arraystretch}{1.2}
\begin{tabular}{@{}>{\centering\arraybackslash}p{2.1cm}|
                >{\centering\arraybackslash}p{1.7cm}|
                >{\centering\arraybackslash}p{1.5cm}|
                >{\centering\arraybackslash}p{1.5cm}@{}}
\toprule
\textbf{Method} & \textbf{CREMAD} & \textbf{K-Sounds} & \textbf{AVE} \\ 
\midrule
Baseline               & 70.88\% & 68.00\% & 70.52\% \\
\midrule
Only-Reshape         & 71.31\% & 67.38\% & 70.31\% \\
Only-TPO               & 75.43\% & 70.94\% & 74.48\% \\
\midrule
Uni-Distill            & 73.86\% & 64.57\% & 54.69\% \\
Vanilla-Reshape           & 69.89\% & 61.29\% & 53.13\% \\
\midrule
BMLR                   & \textbf{79.69\%} & \textbf{71.76\%} & \textbf{76.04\%} \\
\bottomrule
\end{tabular}
\label{tab: ablan}
\end{table}

\subsubsection{Validation of Design Effectiveness}
To systematically evaluate the contribution of each core module and key design to the overall performance, we further conduct comprehensive ablation studies.
The results are shown in Table \ref{tab: ablan}.

Specifically, we first conduct ablation studies on the effectiveness of the two key components. 
In the table, \textbf{Only-Reshape} and \textbf{Only-TPO} denote variants that retain only the Label Reshaping module or only Targeted Parameter Optimization (TPO) in BMLR, respectively. 
The results show that Only-Reshape may lead to degraded performance. This does not indicate that the reshaping mechanism is ineffective, but rather that the dynamic supervision signals introduce considerable instability during training.
In contrast, Only-TPO yields stable performance gains, which is consistent with the findings of DGL \cite{dgl}.
However, it remains clearly inferior to the full BMLR model, suggesting that parameter optimization alone is insufficient to resolve modality imbalance, and that the reshaping mechanism is equally essential. Therefore, \textbf{the two core components of BMLR are tightly coupled and mutually beneficial by design.}

We further evaluate the specific design of the Reshaping module through two variants: \textbf{Uni-Distill} and \textbf{Vanilla-Reshape}.
Uni-Distill removes all imbalance-aware designs and uses only cross-modal prediction logits as supervision signals, highlighting the difference and advantage of our method over conventional cross-modal distillation strategies. 
Vanilla-Reshape preserves the imbalance-aware design but removes the original supervision signal in Equation \ref{eq: reshape_y}. Although this variant produces more balanced supervision signals, the labels become overly smooth and lack clear learning targets, resulting in a substantial performance drop.

Overall, these ablation studies demonstrate that the complete BMLR framework effectively combines label fairness and training stability. It \textbf{provides more appropriate supervision signals while alleviating the optimization difficulty caused by dynamically changing labels}, thereby leading to superior multimodal learning performance.

\begin{table}[t]
\centering
\caption{\textbf{Comparison with Balanced Multimodal Learning Methods.} 
\textbf{Bold} represents the best accuracy in existing methods, while \underline{underline} represents the second-best performance.}
\label{tbl: tri}
\renewcommand{\arraystretch}{1.2}
\begin{tabular}{
@{}
>{\centering\arraybackslash}p{3.5cm}|
>{\centering\arraybackslash}p{1.8cm}|
>{\centering\arraybackslash}p{1.8cm}
@{}}
\toprule
\textbf{Method}  & \textbf{MOSI} & \textbf{MOSEI} \\ 
\midrule
Baseline &  36.16\% & 41.68\%  \\
\midrule
AGM \xsmall{(ICCV 2023)} & \underline{\textbf{37.05\%}} & 41.75\% \\ 
ReconBoost \xsmall{(ICML 2024)} & 33.93\% & \underline{43.69\%}  \\
MMPareto \xsmall{(ICML 2024)} &  36.17\% & 41.92\% \\
OPM \xsmall{(TPAMI 2025)} & 36.16\% & 41.14\%  \\
LFM \xsmall{(NeurIPS 2024)} & 34.48\% & 41.94\%  \\
DGL \xsmall{(ICCV 2025)} & 36.63\% & 42.78\%  \\
\midrule
\textbf{BMLR} \xsmall{(ours)} & \underline{\textbf{37.05\%}} & \textbf{43.71\%} \\
\bottomrule
\end{tabular}
\end{table}

\subsection{Experiments on Advanced Architectures and Complex Multimodal Scenarios}
To further evaluate the effectiveness of our method in more complex multimodal settings, we conduct additional experiments on the trimodal sentiment analysis datasets MOSI and MOSEI.
(1) \textit{To assess compatibility with advanced architectures}, we replace all CNN-based encoders with Transformer-based models. 
Since BMLR operates on the label space, no modification to the method itself is required when changing the encoder architecture.
(2) \textit{To adapt BMLR to the trimodal setting}, we focus on the strongest and weakest modalities during label space reshaping, while keeping the intermediate modality unchanged. 
This design aligns the mapping difficulty of the strong and weak modalities with that of the intermediate modality, enabling BMLR to function effectively in trimodal scenarios.
We further compare our method with several representative BML approaches that can be extended to trimodal settings. 
As shown in Table \ref{tbl: tri}, our method consistently improves performance under both more complex multimodal settings and more advanced architectures, outperforming existing BML methods.

\section{Conclusion}
In this work, we argue that differences in optimization pace across modalities arise from the varying difficulty of mapping between feature and label spaces.
To address this issue, we propose BMLR, the first label-side regulation method for balanced multimodal learning.
By reshaping the label space, BMLR not only aligns learning difficulty across modalities but also enhances inter-class relationship modeling and cross-modal interactions, leading to improved unimodal and multimodal performance.
Extensive experiments on diverse datasets, architectures, and fusion strategies provide consistent empirical evidence for the effectiveness of BMLR, which outperforms existing balanced learning approaches and demonstrates strong generalizability across settings.


\vfill

\end{document}